\pdfoutput=1
\documentclass[11pt]{article}

\usepackage[preprint]{acl}
\usepackage{times}
\usepackage{latexsym}
\usepackage{multirow}
\usepackage{booktabs}
\usepackage{latexsym}
\usepackage{amsmath}
\usepackage{amsthm}
\usepackage{amssymb}
\usepackage{array}
\usepackage{enumitem}
\usepackage{graphicx} 
\usepackage{color}
\usepackage{textcomp}
\usepackage{hyperref}
\usepackage[T1]{fontenc}
\usepackage[utf8]{inputenc}

\usepackage{microtype}

\usepackage{inconsolata}

\usepackage{graphicx}

\def\method{Amphista}
\title{Amphista: Bi-directional Multi-head Decoding for Accelerating LLM Inference}

\author{
    Zeping Li$^{1}$\thanks{Equal contribution.}, Xinlong Yang$^{12*}$, Ziheng Gao$^1$, Ji Liu$^1$, Guanchen Li$^1$, Zhuang Liu$^1$, Dong Li$^1$,\\ \textbf{Jinzhang Peng}$^1$, \textbf{Lu Tian}$^1$, \textbf{Emad Barsoum}$^1$\\
    $^1$ Advanced Micro Devices, Inc.
   $^2$ Peking University\\
    \{\href{mailto:zeping.li@amd.com}{zeping.li},
    \href{mailto:emad.barsoum@amd.com}{emad.barsoum}\}@amd.com
}

\begin{document}
\maketitle      

\begin{abstract}
Large Language Models (LLMs) inherently use autoregressive decoding, which lacks parallelism in inference and results in significantly slow inference speed. While methods such as Medusa constructs parallelized heads, they lack adequate information interaction across different prediction positions. To overcome this limitation, we introduce \textbf{\method{}}, an enhanced speculative decoding framework that builds upon Medusa. Specifically, \method{} models an \textit{Auto-embedding Block} capable of parallel inference, incorporating bi-directional attention to enable interaction between different drafting heads. Additionally, \method{} integrates \textit{Staged Adaptation Layers}, which ensure a seamless transition of semantic information from the target model’s autoregressive inference to the drafting heads' non-autoregressive inference, effectively achieving paradigm shift and feature fusion. Experimental results on Vicuna models using MT-Bench and Spec-Bench demonstrate that \method{} achieves substantial acceleration while maintaining generation quality. On MT-Bench, \method{} delivers up to \textbf{2.75$\times$} speedup over vanilla autoregressive decoding and \textbf{1.40$\times$} over Medusa on Vicuna 33B in wall-clock time.


\end{abstract}



\section{Introduction}

Generative large language models (LLMs) have made remarkable advances in language processing by scaling the transformer decoder block, offering a potential pathway toward Artificial General Intelligence (AGI) \citep{chatgpt, touvron2023llama}. However, the autoregressive nature of next-token prediction and the large parameter size of foundational models result in low inference efficiency, marked by high latency per token and low throughput per second during decoding.

In this context, acceleration during inference has become a burgeoning research area. Speculative decoding \cite{stern2018blockwise, chen2023accelerating} uses a draft model for preliminary multi-step speculative inference and a target model to verify the speculative predictions, emerging as a very promising algorithmic strategy. Notably, by employing a rejection sampling strategy \cite{leviathan2023fast}, the generation quality and accuracy of the speculate-and-verify framework are consistent with those of the target model, making speculative decoding a \textbf{lossless} acceleration framework. Medusa decoding \cite{cai2024medusa} innovatively uses the target model’s last hidden states to implement a multi-heads inference framework. It is widely adopted for its efficient acceleration and simple structure.

\begin{figure}[t]
    \centering
    \includegraphics[width=0.5\textwidth]{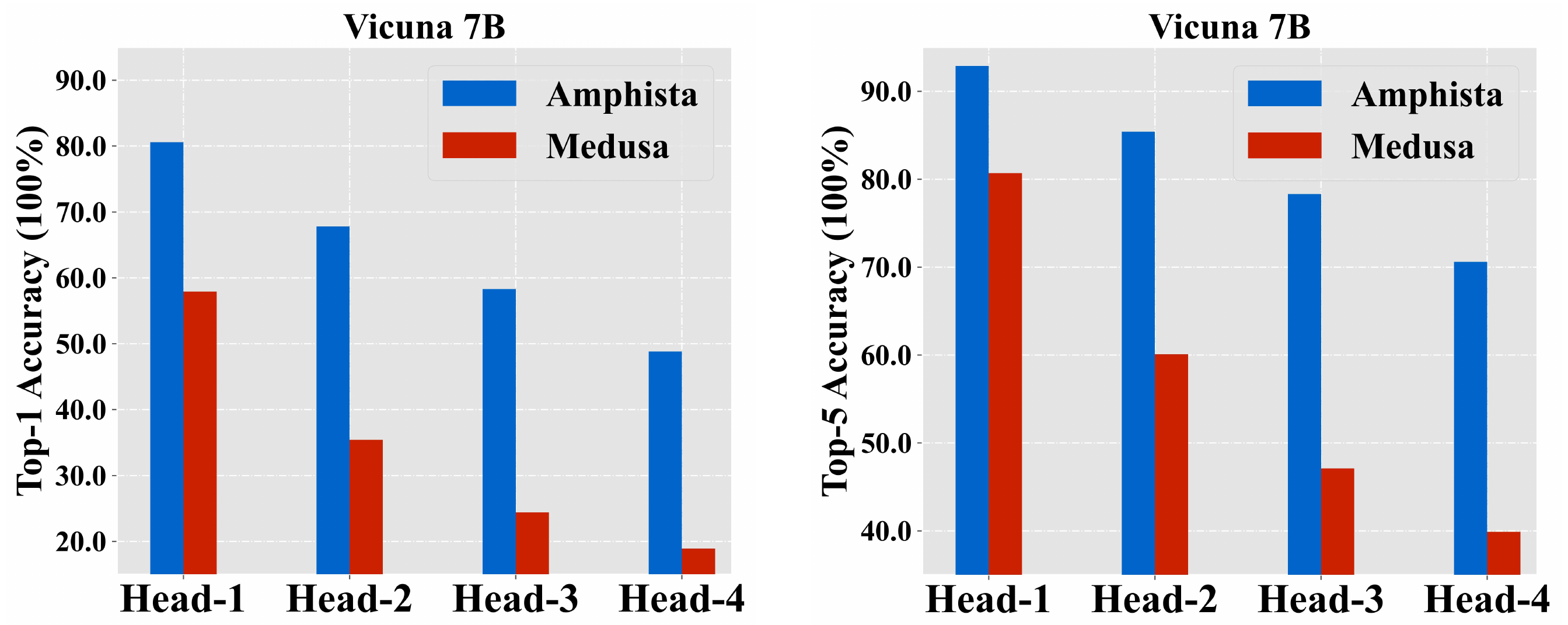}
    \caption{Top-1/5 accuracy for different heads of Medusa and \method{}. We perform testing with randomly sampled 5\% ShareGPT conversation data. \method{} far outperforms Medusa in terms of head accuracy, especially for the latter two heads.}
    \label{fig:top} 
    \vspace{-0.2cm}
\end{figure}

Nonetheless, as illustrated in Figure \ref{fig:top}, we find that the prediction accuracy of separately independent Medusa heads is relatively low, which progressively worsens and adversely impacts acceleration performance in downstream tasks. To mitigate these inaccuracies stemming from the absence of feature dependencies while maintaining parallel inference, we first introduce the Auto-embedding Block, which integrates a bi-directional self-attention mechanism \cite{vaswani2017attention}. This structure not only allows earlier heads to attend to subsequent ones, but more importantly, enables backward heads to leverage information from preceding heads. This enhancement allows drafting heads to better capture contextual information, thereby improving the acceptance rate of their predictions. Moreover, in the multi-step drafting framework, this non-autoregressive structure achieves lower drafting latency compared to an autoregressive approach. 

Additionally, we identify a significant gap between the autoregressive target model and the non-autoregressive draft model in their prediction paradigms. To bridge this discrepancy and further enhance feature representations across different drafting heads, we propose the Staged Adaptation Layers. These layers serve as an intermediary module to facilitate feature integration and transformation between the target model and draft heads. Once adopted, semantically enriched features are passed through MLP activations and fed into the auto-embedding block. This enhances the bi-directional attention mechanism's ability to fuse features across heads, ultimately boosting acceptance rates and reducing wall-clock time.

Lastly, to further align the draft model with the target model with minimal computational cost, we incorporate the sampled token from the target model’s latest prediction into the staged adaptation layers. This critically integrated information harmonizes \method{} with the target model, yielding a significant improvement in performance.

On MT-Bench, \method{} achieves up to \textbf{2.75$\times$} speedup over vanilla autoregressive decoding and \textbf{1.40$\times$} over Medusa on Vicuna 33B, as consistently high accuracy (see Figure \ref{fig:top}). To summarize, our contributions are as follows:
\begin{itemize}[leftmargin=*]
  \item We present \method{}, a non-autoregressive and innovatively cost-efficient inference acceleration framework, built upon the foundational principles of Medusa decoding.
  \item We introduce the Auto-embedding Block, which enables bi-directional interaction among different heads by facilitating collaborative information exchange during the drafting phase. Additionally, the Staged Adaptation Layers are introduced to bridge the gap between autoregressive and non-autoregressive token prediction through a two-stage adaptation process. Finally, the integration of a sampled token from the target model further aligns the draft and target models with minimal computational overhead.
  \item We perform comprehensive evaluations on a diverse set of foundational models. The results show that \method{} consistently outperforms Medusa in terms of both acceptance rate and speed-up, across various generation tasks.
\end{itemize}

\section{Preliminaries}

\begin{figure*}[t]
    \centering
    \includegraphics[width=0.95\textwidth]{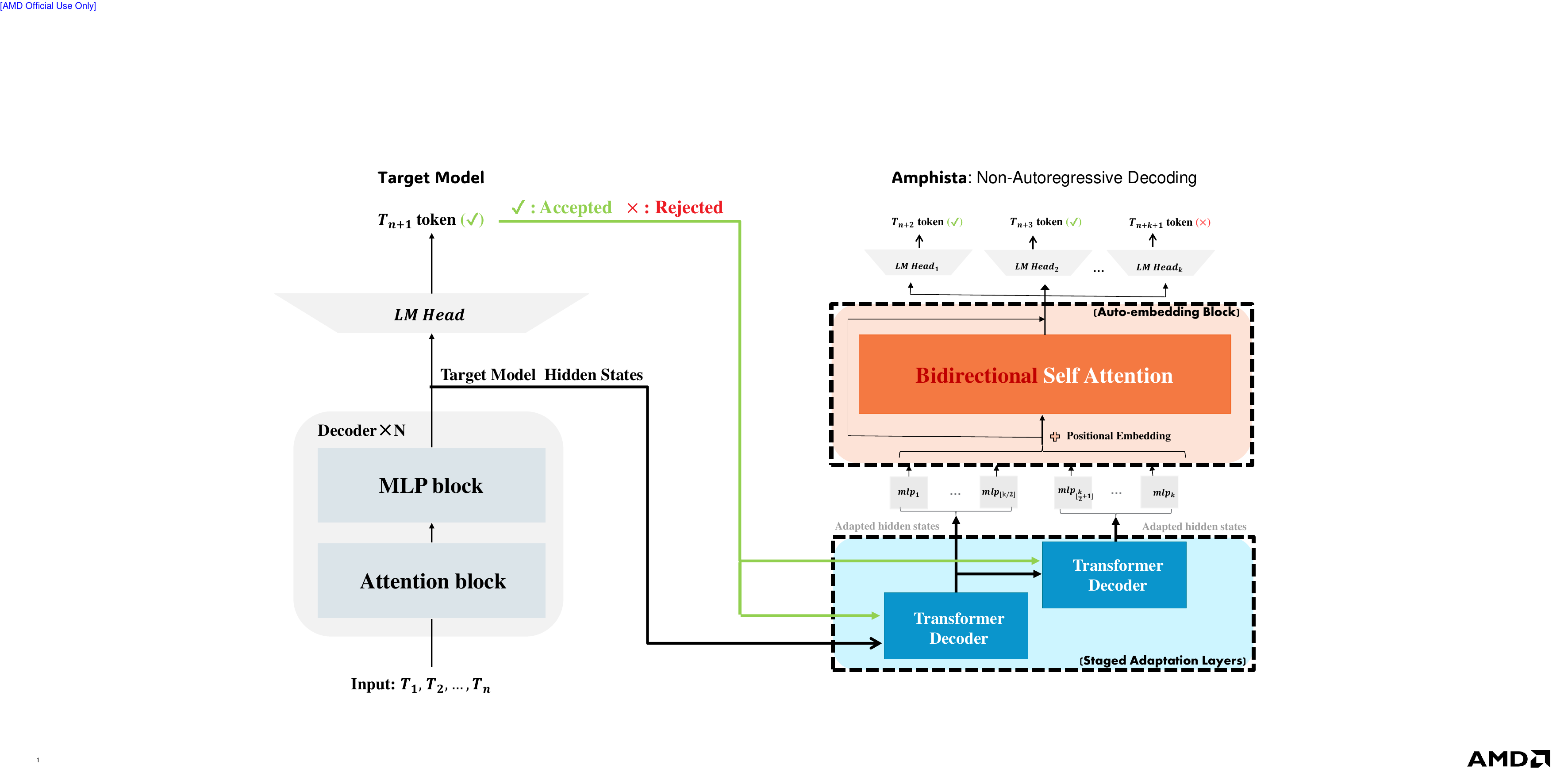}
    \caption{\textbf{The Framework of \method{} Decoding.} Our method improve Medusa in two folds: (1) We introduce staged adaptation layers, consisting of a group of causal Transformer Decoder layers built upon the target model, to adapt the target model's hidden states and the sampled token in two stages. This module ensures that the adapted features contain richer contextual information, supporting multiple-token predictions rather than focusing solely on the immediate next-token prediction. (2) We introduce an auto-embedding block, which is a bi-directional Transformer Encoder module with positional encoding. This block allows each head to attend to others, fostering cooperative predictions and thereby enhancing the speculative accuracy during the drafting stage.}
    \label{fig:main} 
\end{figure*}

In this section, we introduce some preliminary
background related to our work as follows:

\noindent\textbf{Speculative Decoding.} Speculative decoding has been successfully applied to LLM decoding algorithm recently \cite{leviathan2023fast, chen2023accelerating}. The core idea is to leverage a small, lower-quality model (draft model) together with a large, higher-quality model (target model) to accelerate token generation. Concretely, in each decoding step, the algorithm first uses the draft model to autoregressively generate a sequence of future tokens. These drafted tokens are then verified by the target model in a single forward pass. During the verification process, a certain strategy is applied to determine which tokens are accepted by the target model and which are rejected and discarded. Previous work \cite{leviathan2023fast} has theoretically and empirically demonstrated that the token output distribution of speculative decoding is consistent with the autoregressive generation of original target model, but with fewer decoding steps, thus enhancing generation efficiency.

\noindent\textbf{Medusa Decoding.} Medusa Decoding \cite{cai2024medusa} represents an efficient speculative decoding algorithm based on the draft-and-verify principle, inheriting principles from blockwise parallel decoding \cite{stern2018blockwise}. Specifically, Medusa integrates independent MLP layers, called drafting heads, with the target model to form a unified architecture. In each decoding step, the target model's \texttt{lm\_head} samples the next token, while the next-i MLP heads predict tokens at subsequent positions. These drafted tokens are then verified by the target model’s forward pass to decide their acceptance. By leveraging lightweight MLP layers, Medusa strikes an effective balance between computational efficiency and prediction accuracy, leading to substantial acceleration. Hydra \cite{ankner2024hydra}, which is a subsequent state-of-the-art optimization based on Medusa, transforms the independent MLP heads into sequentially dependent MLP heads, further enhancing the predictive accuracy. 

\noindent\textbf{Tree Attention.} 
Tree attention \cite{miao2024specinfer, cai2024medusa} enables parallel computation of attention scores for multiple draft candidates. Medusa uses a tree causal mask, allowing each node to attend only to its ancestors, efficiently processing multiple candidate sequences simultaneously (see Appendix \ref{app:draft} for details).


\section{\method{}}
The overview of \method{} is shown in Figure \ref{fig:main}. Building its pipeline upon target model, \method{} contains two main modules: (1) Staged Adaptation Layers. They are causal Transformer Decoder layers that adapt the target model's hidden states and sampled token embedding in two stages, each focusing on different drafting positions. This adaptation process results in hidden states that are enhanced with position-aware contextual information, improving overall prediction accuracy, especially for the latter steps. (2) Auto-embedding Block. It is a Transformer Encoder module that conducts \textit{bi-directional} self-attention computations among the representations of different draft heads, allowing each head can be attended by the others. This facilitates collaborative prediction among these heads, thereby improving overall prediction accuracy. 

\subsection{Staged Adaptation Layers}
Figure \ref{fig:main} demonstrates the relevant details of our staged adaptation layers. Although target model's hidden states contain semantically rich information, there are still differences in the representation requirements between the target model and the draft heads. Specifically, the hidden states of the target model are trained only for predicting the next token, while draft heads need more contextual and positon-aware hidden states to perform multi-step speculation. To address this problem, Medusa-2 applies LoRA \cite{hu2021lora} for joint training of the target model and draft heads, which may compromise the generality on downstream tasks. Hydra employs a single prefix layer for all positions, lacking targeted adaptation for different positions. We propose an effective adaptation method by incorporating two adaptation layers to transform and adapt the strong semantic information from the target model in stages. Specifically, given the hidden states $h_t$ at position t from the target model's final layer and the embedding of the token $e_{t+1}$ sampled from $h_t$, we use the two adaptation layers to transform them in stages as below:
\begin{equation}
\begin{aligned}
    & \quad h^{1}_t = SAL^{1}(fc^1([h_t; e_{t+1}]), kv^{1}_{1:t-1}), \\
    & \quad h^{2}_t = SAL^{2}(fc^2([h^{1}_t; e_{t+1}]), kv^{2}_{1:t-1}).
\end{aligned}
\end{equation}

$SAL^{1}$ stands for the Stage-one Adaptation Layer that adapts target model hidden states and sampled token embedding, while $SAL^{2}$ stands for the Stage-two Adaptation Layer that adapts $SAL^{1}$'s output hidden states as well as the sampled token embedding. The function $fc^1$ and $fc^2$ are fully connected layers employed to transform features derived from the concatenation of hidden states and token embeddings. The terms ${kv}^{1}_{1:t-1}$ and ${kv}^{2}_{1:t-1}$ represent the key-value caches for each adaptation layer. Subsequently, adapted hidden states \( h^{1}_{t} \) and \( h^{2}_{t} \) are fed into the first and second halves of the drafting heads respectively, ensuring that each adaptation layer focuses on adapting target model's semantic representations in specific future locations.

\subsection{Auto-embedding Block}\label{auto-embed}
Figure \ref{fig:main} shows the detailed design of our Auto-embedding Block. Given a set of $K$ drafting MLP heads, $\text{MLP}_k$ head is tasked with predicting the token in the $(t+k+1)$-th position. Upon obtaining adapted hidden states \( h^{1}_{t} \) and \( h^{2}_{t} \) from the first and second staged adaptation layers, we first utilize the MLP layers to project them into more position-aware and semantically rich hidden states:
\begin{equation}\label{mlp}
\begin{aligned}
h^{'}_k &= \text{MLP}_k(h^{1}_{t}), &\, k = 1, 2, \ldots, \lfloor K/2 \rfloor, \\
h^{'}_k &= \text{MLP}_k(h^{2}_{t}), &\, k = \lfloor K/2 \rfloor +1, \ldots, K,
\end{aligned}
\end{equation}
where \( \text{MLP}_i \in \mathbb{R}^{d \times d} \), and \( d \) is the dimension of the target model hidden states. We then concatenate these \( K \) hidden states along the \texttt{seq\_len} dimension:
\begin{equation}\label{eq:concat}
H' = \operatorname{concat}([h'_1, h'_2, h'_3, \dots, h'_K]),
\end{equation}
where \( H' \in \mathbb{R}^{K \times d} \). In order to further enhance the relative positional information among different heads, we introduce additional positional encodings. Specifically, we introduce a learnable positional embedding \( PE \in \mathbb{R}^{K \times d} \), and the position-encoded hidden states $H_p$ are expressed as:
\begin{equation}\label{eq:posi}
H_p = H' + PE.
\end{equation}

Finally, we employ an effective and efficient bi-directional self-attention module to enable mutual awareness among the drafting heads and use additional learnable lm-head to sample the top-\( k \) draft tokens in each position:
\begin{equation}
    attn_o  = \text{Self-Attention}(H_p),
\end{equation}
\begin{equation}
    d\_logits_k  = \textstyle \text{LM-Head}_{k}(\smash{attn_o[k]}), \quad k=1,\ldots,K.
\end{equation}



In the end, these draft tokens are organized into a draft tree and then verified by the LLM through tree attention. Unlike the independent heads in Medusa and the sequentially dependent heads in Hydra, \method{} adopts bi-directionally dependent heads. This approach enhances overall prediction accuracy while maintaining a non-autoregressive mechanism, potentially reducing the substantial computation overhead associated with sequential calculations (i.e., autoregressive manner).


\begin{table*}[t]
\centering
\caption{The speed-up comparison on MT-Bench and Spec-Bench between different methods under \textbf{greedy sampling} setting (Temperature = 0). We regard the speed-up of vanilla autoregressive decoding as 1.00$\times$.}
\label{tab:main1}
    \resizebox{0.9\textwidth}{!}{
\begin{tabular}{ll|c|cccccc}
\toprule
\multirow{2}{*}{Model Size} & \multirow{2}{*}{Method} & \multirow{2}{*}{MT-Bench} & \multicolumn{5}{c}{Spec-Bench} & \multirow{2}{*}{Avg} \\ \cline{4-8}
 &  &  & Translation & Summarization & QA & Math & RAG &  \\ \midrule
\multirow{5}{*}{7B} & Vanilla & 1.00$\times$ & 1.00$\times$ & 1.00$\times$ & 1.00$\times$ & 1.00$\times$ & 1.00$\times$ & 1.00$\times$  \\
 & Spec-decoding & 1.62$\times$ & 1.11$\times$ &1.66$\times$  & 1.46$\times$ & 1.45$\times$ & 1.61$\times$ & 1.45$\times$ \\
  & Lookahead & 1.44$\times$ & 1.15$\times$ & 1.26$\times$ & 1.25$\times$ & 1.56$\times$& 1.13$\times$ & 1.27$\times$ \\
 & Medusa & 1.87$\times$ & 1.42$\times$ & 1.42$\times$ &1.50$\times$ & 1.74$\times$ & 1.39$\times$  & 1.50$\times$\\
 & Hydra++ & 2.37$\times$ & 1.92$\times$ & 1.80$\times$ & \textbf{1.94$\times$} & 2.43$\times$ & 2.04$\times$ & 2.03$\times$ \\
 & \textbf{\method{} (ours)} & \textbf{2.44$\times$} & \textbf{1.96$\times$} & \textbf{2.11$\times$} & \textbf{1.94$\times$} & \textbf{2.45$\times$} & \textbf{2.20$\times$} & \textbf{2.13$\times$} \\ \midrule
\multirow{5}{*}{13B} & Vanilla  & 1.00$\times$ & 1.00$\times$ & 1.00$\times$ & 1.00$\times$ & 1.00$\times$ & 1.00$\times$ & 1.00$\times$  \\
 & Spec-decoding & 1.66$\times$ & 1.17$\times$ & 1.75$\times$ & 1.44$\times$ & 1.59$\times$ &  1.73$\times$ &  1.53$\times$\\
  & Lookahead & 1.34$\times$ & 1.08$\times$ & 1.23$\times$ & 1.15$\times$ & 1.51$\times$& 1.15$\times$ & 1.22$\times$ \\
 & Medusa & 1.85$\times$ & 1.55$\times$ & 1.55$\times$ & 1.53$\times$ & 1.88$\times$ & 1.51$\times$ & 1.60$\times$ \\
 & Hydra++ & 2.34$\times$ & 1.75$\times$ & 1.85$\times$ & 1.85$\times$ & 2.31$\times$ & 1.86$\times$ & 1.92$\times$\\
 & \textbf{\method{} (ours)} & \textbf{2.49$\times$} & \textbf{1.88$\times$}  & \textbf{2.14$\times$} & \textbf{1.88$\times$} & \textbf{2.41$\times$} & \textbf{2.04$\times$} & \textbf{2.07$\times$} \\ \midrule
\multirow{5}{*}{33B}  & Vanilla & 1.00$\times$ & 1.00$\times$ & 1.00$\times$ & 1.00$\times$ & 1.00$\times$ & 1.00$\times$ & 1.00$\times$  \\
 & Spec-decoding &1.73$\times$  & 1.28$\times$ & 1.76$\times$ & 1.54$\times$ & 1.71$\times$ &  1.69$\times$& 1.60$\times$ \\
  & Lookahead & 1.32$\times$ & 1.09$\times$ & 1.21$\times$ & 1.16$\times$ & 1.55$\times$& 1.16$\times$ & 1.24$\times$ \\
 & Medusa & 1.97$\times$ & 1.72$\times$ & 1.62$\times$ & 1.66$\times$ & 2.06$\times$ & 1.61$\times$ & 1.73$\times$ \\
 & Hydra++ & 2.54$\times$ & 1.93$\times$ & 2.10$\times$ & 2.04$\times$ & 2.63$\times$ & 2.17$\times$ & 2.17$\times$ \\
 & \textbf{\method{} (ours)} & \textbf{2.75$\times$} & \textbf{2.11$\times$} & \textbf{2.49$\times$} & \textbf{2.12$\times$} & \textbf{2.83$\times$} & \textbf{2.44$\times$}  & \textbf{2.40$\times$} \\ \bottomrule
\end{tabular}}
\end{table*}

\subsection{Training Objective}
Our loss function integrates two components to achieve a dual objective. First, we employ a Cross-Entropy (CE) loss between the logits of \method{} and those of the target model to align their output token distributions. Second, we utilize a language modeling (LM) loss to quantify the discrepancy between \method{}'s outputs and the ground truth tokens. This approach enables \method{} not only to emulate the target model but also to assimilate predictive capabilities from the real corpus.
\begin{equation}
\mathcal{L}_{\text{\method{}}} = \lambda_1 \, \mathcal{L}_{\text{alignment}} + \lambda_2 \, \mathcal{L}_{\text{lm}},
\end{equation}
\begin{equation}
\mathcal{L}_{\text{alignment}} = \text{CE}(d\_logits_i, logits_{T_{t+1+i}}),
\end{equation}
\begin{equation}
\mathcal{L}_{\text{lm}} = \text{CE}(d\_logits_i, gt_{t+1+i}).
\end{equation}

Note that \( d\_logits_i \) and \(logits_{T_{t+1+i}}\) are the logits from \method{} and the target model for token $T_{t+1+i}$, while \( gt_{t+1+i} \) represent the ground truth labels of token $T_{t+1+i}$. The terms \( \lambda_1 \) and \( \lambda_2 \) are weighting factors for the two objectives.

\section{Experiments}

\begin{table*}[t]
\centering
\caption{The speed-up comparison on MT-Bench and Spec-bench between different methods under \textbf{random sampling} setting (Temperature = 0.7). We regard the speed-up of vanilla autoregressive decoding as 1.00$\times$.}
\label{tab:main2}
    \resizebox{0.9\textwidth}{!}{
\begin{tabular}{ll|c|cccccc}
\toprule
\multirow{2}{*}{Model Size} & \multirow{2}{*}{Method} & \multirow{2}{*}{MT-Bench} & \multicolumn{5}{c}{Spec-Bench} & \multirow{2}{*}{Avg} \\ \cline{4-8}
 &  &  & Translation & Summarization & QA & Math & RAG &  \\ \midrule
\multirow{5}{*}{7B} & Vanilla & 1.00$\times$ & 1.00$\times$ & 1.00$\times$ & 1.00$\times$ & 1.00$\times$ & 1.00$\times$ & 1.00$\times$  \\
 & Spec-decoding & 1.39$\times$ & 1.02$\times$ & 1.41$\times$ & 1.24$\times$ & 1.32$\times$& 1.43$\times$ & 1.28$\times$ \\
 & Lookahead & 1.28$\times$ & 1.05$\times$ & 1.21$\times$ & 1.12$\times$ &1.25$\times$& 1.14$\times$ & 1.16$\times$ \\
 & Medusa & 1.86$\times$ & 1.51$\times$ & 1.47$\times$ & 1.57$\times$ & 1.89$\times$ & 1.43$\times$ & 1.57$\times$ \\
 & Hydra++ & 2.35$\times$  & \textbf{1.81$\times$} & 1.81$\times$ & \textbf{1.97$\times$} & 2.41$\times$ & 1.74$\times$  & 1.95$\times$ \\
 & \textbf{\method{} (ours)} & \textbf{2.37$\times$} & \textbf{1.81$\times$} & \textbf{1.92$\times$} & 1.96$\times$ & \textbf{2.43$\times$} & \textbf{1.79$\times$} & \textbf{1.99$\times$} \\ \midrule
\multirow{5}{*}{13B} & Vanilla  & 1.00$\times$ & 1.00$\times$ & 1.00$\times$ & 1.00$\times$ & 1.00$\times$ & 1.00$\times$ & 1.00$\times$  \\
 & Spec-decoding & 1.52$\times$ & 1.08$\times$ & 1.57$\times$ & 1.33$\times$ & 1.42$\times$ & 1.46$\times$ & 1.37$\times$\\
  & Lookahead & 1.30$\times$ & 1.07$\times$ & 1.19$\times$ & 1.15$\times$ & 1.38$\times$& 1.14$\times$ & 1.19$\times$ \\
 & Medusa & 2.01$\times$ & 1.65$\times$ &1.62$\times$  & 1.71$\times$ & 2.01$\times$ & 1.57$\times$ & 1.71$\times$ \\
 & Hydra++ & 2.57$\times$ &1.90$\times$  & 1.99$\times$ & 2.12$\times$ & 2.56$\times$ & 2.04$\times$ & 2.12$\times$\\
 & \textbf{\method{} (ours)} & \textbf{2.65$\times$} & \textbf{1.93$\times$}  & \textbf{2.16$\times$} & \textbf{2.17$\times$} & \textbf{2.64$\times$} & \textbf{2.15$\times$} & \textbf{2.22$\times$} \\ \midrule
\multirow{5}{*}{33B}  & Vanilla & 1.00$\times$ & 1.00$\times$ & 1.00$\times$ & 1.00$\times$ & 1.00$\times$ & 1.00$\times$ & 1.00$\times$  \\
 & Spec-decoding & 1.58$\times$ & 1.21$\times$ & 1.62$\times$ & 1.48$\times$ & 1.59$\times$ & 1.54$\times$& 1.48$\times$ \\
  & Lookahead & 1.29$\times$ & 1.04$\times$ & 1.18$\times$ & 1.15$\times$ & 1.52$\times$& 1.14$\times$ & 1.21$\times$ \\
 & Medusa & 2.06$\times$ &1.71$\times$  & 1.79$\times$ & 1.76$\times$  & 2.10$\times$ & 1.79$\times$  & 1.83$\times$ \\
 & Hydra++ & 2.74$\times$ & 2.01$\times$ & 2.24$\times$ & 2.24$\times$  & 2.82$\times$ & 2.26$\times$ & 2.31$\times$ \\
 & \textbf{\method{} (ours)} & \textbf{2.85$\times$} & \textbf{2.05$\times$}  & \textbf{2.51$\times$} & \textbf{2.29$\times$} & \textbf{2.90$\times$} & \textbf{2.39$\times$} & \textbf{2.43$\times$} \\ \bottomrule
\end{tabular}}
\end{table*}

\subsection{Experimental Settings}
\noindent\textbf{Models and Baselines.} Following \cite{cai2024medusa,li2024eagle, ankner2024hydra}, we use Vicuna family of models \cite{zheng2024judging} as our target model. Specifically, we implement our method on Vicuna 7, 13, and 33B models with four drafting heads. As for compared baseline methods, we choose original Speculative Decoding, Lookahead \cite{fu2024break}, Medusa \cite{cai2024medusa} and Hydra \cite{ankner2024hydra} for comparison.

\noindent\textbf{Training and Datasets.} For the training stage, again following \cite{cai2024medusa, ankner2024hydra}, we use ShareGPT \footnote{ShareGPT. 2023. \href{https://huggingface.co/datasets/Aeala/ShareGPT_Vicuna_unfiltered}{https://huggingface.co/datasets/Aeala/\\ShareGPT\_Vicuna\_unfiltered}} dataset to fine-tune our proposed module while keeping target model frozen. Training is conducted using HuggingFace Trainer, which we employ with AdamW optimizer ($\beta_{1}$=0.9, $\beta_{2}$=0.999) and a cosine learning rate schedule with warmup strategy, the initial learning rate is set to 1e-3 and we train 4 epochs. At the evaluation stage, we use MT-Bench \cite{zheng2024judging} and Spec-Bench \cite{xia2024unlocking} as our benchmark. MT-Bench is an open source multi-turn conversation benchmark. Spec-Bench is a well-acknowledged and comprehensive benchmark designed for assessing speculative decoding methods across diverse application scenarios.

\noindent\textbf{Metrics.} Following previous speculative decoding work, we choose tokens/s and tokens/step as our main metrics. Tokens/step measures the average token length accepted per forward pass of the target LLM. Tokens/s represents the overall throughput of the acceleration algorithm, which is influenced by both the prediction accuracy of the speculator and the drafting latency of the speculator.

\subsection{Evaluation of \method{}}
\method{} builds on Medusa to support parallel decoding, distinctly diverging from auto-regression drafting methods. Thus, the representative work of parallel drafting (i.e., Lookahead), and the state-of-the-art work based on Medusa (i.e., Hydra), have been chosen as a competitive baseline method for comparison. Specifically, Hydra's best-performing model (i.e., Hydra++) is used for fair evaluation and vicuna-68m \cite{yang2024multicandidate} is used as draft model for the vanilla speculative decoding method. We conduct all the experiments on A100 40G GPUs, and all the experimental settings are kept the same for fair comparison.

Table \ref{tab:main1} and Table \ref{tab:main2} present the speed-up metrics compared on MT-Bench and Spec-Bench under greedy and random sampling settings (see \ref{app:moreex} for more experiment results). Overall, \method{} demonstrates significant performance superiority over Medusa and surpasses Hydra's best results by a considerable margin across a variety of generation tasks, and also greatly exceeding the speed-up achieved by vanilla speculative decoding. In detail, \method{} achieves a \textbf{2.44$\times$} - \textbf{2.75$\times$} speed-up on MT-Bench and \textbf{2.13$\times$} - \textbf{2.40$\times$} speed-up on Spec-Bench under greedy decoding setting. Similarly, under random sampling setting, \method{} achieves a \textbf{2.37$\times$} - \textbf{2.85$\times$} speed-up and \textbf{1.99$\times$} - \textbf{2.43$\times$} speed-up on MT-Bench and Spec-Bench with different target model sizes. These robust results demonstrate that enhancing non-autoregressive drafting can surpass autoregressive drafting in terms of speed-up, highlighting the efficiency of our \method{} architecture. During the drafting stage, all computations in non-autoregressive modeling (i.e., \method{}) can be processed in parallel, better leveraging the parallel computing capabilities of modern GPU accelerators. This leads to a more optimal trade-off between drafting acceptance rate and drafting latency.

Moreover, \method{} exhibits a discernible upward trend in speed-up when employed on larger target models. This can be attributed to \method{}'s cost-efficient non-autoregressive modeling and effective transformation of semantic information from the target model. \method{} allows for appropriate increases in accepted token length without introducing excessive additional inference costs. For more exploration on the performance potential of \method{}, please refer to \ref{app:opt}. For more exploration on the parameter complexity optimization, please refer to \ref{app:plex}.

\begin{figure}[t]
    \centering
    \includegraphics[width=0.5\textwidth]{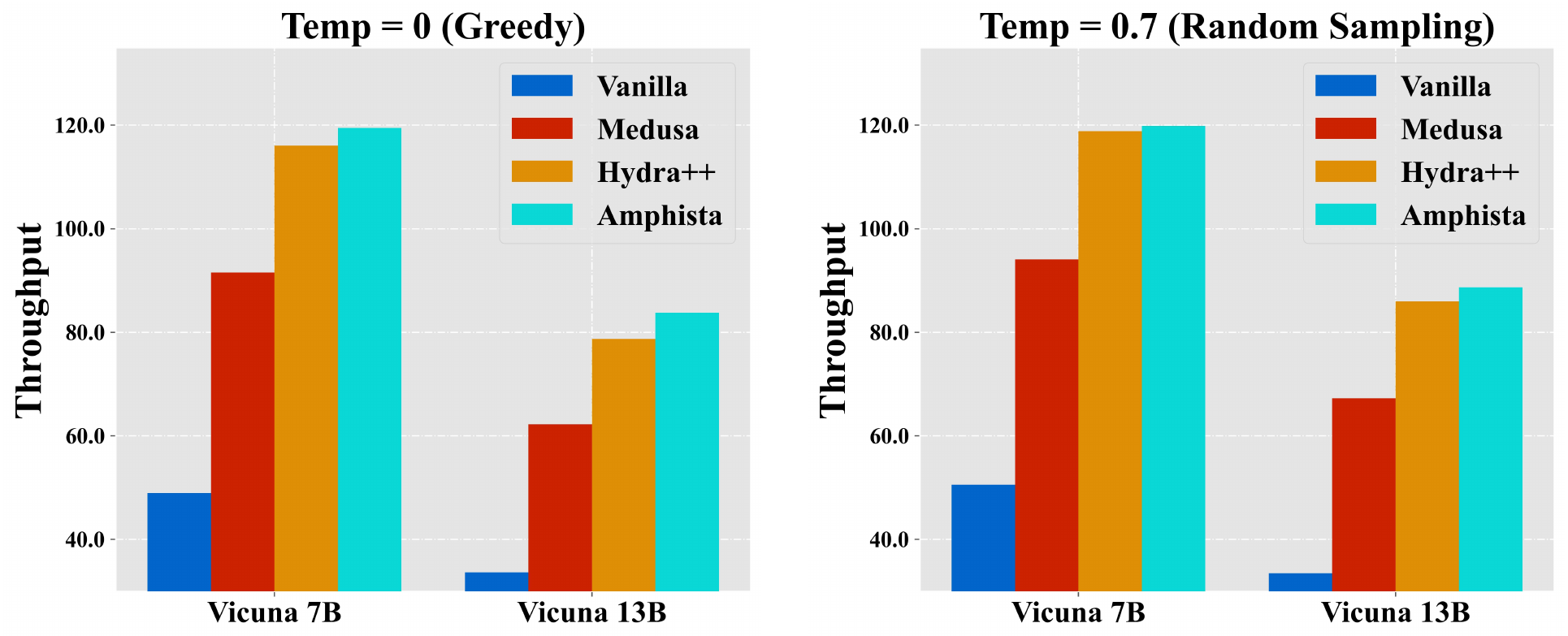}
    \caption{Throughput (tokens/s) on MT-Bench with different target model sizes and temperatures.}
    \label{fig:throughput} 
\end{figure}

Last but not least, we further provide the actual throughput of different methods on MT-Bench with a batch size of 1. As depicted in Figure \ref{fig:throughput}, \method{} achieves an actual throughput of approximately 120 tokens/s with a 7B target model and about 80 tokens/s with a 13B target model under both temperature settings. This performance surpasses that of Medusa and Hydra, underscoring \method{}'s advantages in practical deployment.

\begin{table}
\centering
\caption{Results on CNN/DM and XSUM with different temperatures, AR means Auto-Regressive decoding.}
\label{tab:quality}
    \resizebox{0.5\textwidth}{!}{
\begin{tabular}{@{}l|cccccc@{}}
\toprule
Benchmark  & Temp & Method & ROUGE-1  & ROUGE-2 & ROUGE-L & Speed-up \\ \midrule
 
&    & AR    & 18.74    & 8.44 &  12.59   & 1.00$\times$ \\
 
& \multirow{-2}{*}{0.0} & Amphista                             &18.70  &8.44 & 12.59  &2.15$\times$          \\ \cmidrule(l){2-7} 
 
&   & AR   & 17.92  & 7.65   & 11.93   & 1.00$\times$                           \\
\multirow{-4}{*}{CNN/DM}  & \multirow{-2}{*}{0.7} & \method{}  & 17.91    & 7.65    & 11.92   &2.31$\times$                          \\ \midrule
&       & AR      & 17.32     &  5.05   & 12.16   & 1.00$\times$       \\
 & \multirow{-2}{*}{{ 0.0}} & {Amphista}               & 17.30     &  5.05   & 12.15      &2.25$\times$                  \\ \cmidrule(l){2-7} 
&       & AR    & 15.99     & 4.44    &11.42   &  1.00$\times$        \\
\multirow{-4}{*}{XSUM} & \multirow{-2}{*}{0.7} & Amphista & 15.96 & 4.43 & 11.40     &  2.10$\times$                                                       \\ \bottomrule
\end{tabular}}
\end{table}

\begin{table*}[t]
\centering
\caption{Ablation experiments of different model variants on MT-Bench and Spec-Bench, with the target model being Vicuna 7B and the evaluation metric being \textbf{speed-up}. Medusa can be considered as \method{} w/o any added modules, and Hydra can be seen as Medusa w/ sequential dependency heads.}
\label{tab:abla1}
    \resizebox{0.9\textwidth}{!}{
\begin{tabular}{l|c|cccccc}
\toprule
\multirow{2}{*}{Method Variants} & \multirow{2}{*}{MT-Bench} & \multicolumn{5}{c}{Spec-Bench} & \multirow{2}{*}{Avg} \\ \cline{3-7}
 &  & Translation & Summary & QA & Math & RAG &  \\ \midrule
 Medusa & 1.86$\times$ & 1.51$\times$ & 1.47$\times$ & 1.57$\times$ &1.89$\times$  & 1.43$\times$ & 1.57$\times$ \\
 Hydra++ & 2.37$\times$ & 1.92$\times$ & 1.80$\times$ & 1.94$\times$ &  2.43$\times$& 2.04$\times$ & 2.03$\times$ \\
\method{} w/o Auto-embedding & 2.30$\times$ & 1.82$\times$ & 2.00$\times$ & 1.81$\times$ & 2.25$\times$  & 1.99$\times$ & 1.97$\times$ \\
\method{} w/o Position-Encoding & 2.42$\times$ & 1.96$\times$ & 2.08$\times$ & 1.92$\times$  & 2.42$\times$ & 2.18$\times$ & 2.11$\times$ \\
\method{} w/o Staged-Adaptation & 2.14$\times$ & 1.85$\times$  & 1.75$\times$ & 1.78$\times$ & 2.10$\times$ & 1.91$\times$ & 1.88$\times$ \\
\method{} w/ One-Adaptation-Layer & 2.31$\times$ & 1.90$\times$ & 1.99$\times$ & 1.83$\times$ & 2.35$\times$ & 2.14$\times$ & 2.04$\times$ \\
\method{} w/o Sampled-Token & 2.25$\times$ & 1.88$\times$  & 1.80$\times$ & 1.81$\times$ & 2.26$\times$ & 2.01$\times$ & 1.95$\times$ \\
\textbf{\method{} (ours)} & 2.44$\times$ & 1.96$\times$ & 2.11$\times$ & 1.94$\times$ & 2.45$\times$ & 2.20$\times$ & \textbf{2.13$\times$} \\ \bottomrule
\end{tabular}}
\end{table*}

\begin{table*}[t]
\centering
\caption{Ablation experiments of different model variants on MT-Bench and Spec-Bench, with the target model being Vicuna 7B and evaluation metric being \textbf{average accepted length}. Medusa can be considered as \method{} w/o any added modules, and Hydra can be seen as Medusa w/ sequential dependency heads.}
\label{tab:abla2}
    \resizebox{0.9\textwidth}{!}{
\begin{tabular}{l|c|cccccc}
\toprule
\multirow{2}{*}{Method Variants} & \multirow{2}{*}{MT-Bench} & \multicolumn{5}{c}{Spec-Bench} & \multirow{2}{*}{Avg} \\ \cline{3-7}
 &  & Translation & Summary & QA & Math & RAG &  \\ \midrule
 Medusa & 2.52 & 2.12 & 2.01 & 2.05 & 2.48 & 2.09 & 2.15 \\
 Hydra++ & 3.58 & 2.80 & 2.70 & 2.91 & 3.61 & 2.90 & \textbf{2.98} \\
\method{} w/o Auto-embedding & 3.16  & 2.41 & 2.66 & 2.40 & 3.11 & 2.49 & 2.60 \\
\method{} w/o Position-Encoding & 3.47 & 2.61 & 2.90 & 2.78 & 3.47 & 2.91  & 2.93 \\
\method{} w/o Staged-Adaptation & 2.91 & 2.42 & 2.24 &2.30  & 2.85 & 2.38 & 2.43  \\
\method{} w/ One-Adaptation-Layer & 3.36 & 2.49 & 2.68 & 2.71 & 3.37 & 2.75 & 2.80 \\
\method{} w/o Sampled-Token & 3.11 & 2.43  & 2.48 & 2.45 & 3.15  & 2.55 & 2.61 \\
\textbf{\method{} (ours)} & 3.50 & 2.62 & 3.01 & 2.80  & 3.50 & 2.96 & \textbf{2.98} \\ \bottomrule
\end{tabular}}
\end{table*}

\subsection{Generation Quality of \method{}}
We perform evaluation on XSUM \citep{Narayan2018DontGM} and CNN/DM \citep{see-etal-2017-get} to validate the generation quality of our \method{} (more results can be found in appendix \ref{app:xsum}). From the ROUGE-1/2/L scores \citep{lin2004rouge} in Table \ref{tab:quality}, we can find that \method{} can reserve the output distribution quality while achieving 2.10$\times$-2.31$\times$ speed-up compared with vanilla auto-regressive decoding. 

\subsection{Multi-Batching Exploration}
In this section, we evaluate the speed-up of \method{} in multi-batch scenarios (batch size > 1). For varying sentence lengths within a batch, we use padding to align them and always track the position of the last valid token for each sentence. The experimental results, presented in Table \ref{tab:batch}, are based on randomly sampled prompts from MT-Bench to generate various batch sizes. Generally, as batch size increases, the GPU's idle computational resources gradually decrease, resulting in a reduced speed-up. Additionally, despite the additional computational overhead from different multi-batching strategies, \method{} consistently achieves around 2$\times$ speed-up using the simplest padding method, demonstrating its acceleration advantage in multi-batch settings.


\begin{table}[h]
\centering
\caption{Speed-up evaluation of \method{} on MT-Bench with batch size $> 1$.}
\label{tab:batch}
    \resizebox{0.48\textwidth}{!}{
\begin{tabular}{@{}cccccc@{}}
\toprule
Batch Size & 1    & 2     & 4     & 6     & 8     \\ \midrule
Speed-up   & $2.32\times$ & $2.30\times$ & $2.23\times$ & $2.11\times$ & $2.06\times$ \\ \bottomrule
\end{tabular}}
\end{table}

\subsection{Ablation Study}
 Diverging from other approaches based on speculative sampling and Medusa, \method{}'s main insight lies in adapting transformation through Staged Adaptation Layers and enhancing integration via the non-autoregressive Auto-embedding Block. These approaches strengthen semantic information derived from the target model. In this section, we conduct comprehensive ablation experiments based on the vicuna 7B model to validate the effectiveness of each proposed module in our \method{}. Specifically, we conduct five model variants as follows: (1) \textbf{\method{} w/o Auto-embedding} which means removing the Auto-embedding Block. (2) \textbf{\method{} w/o Position-Encoding} which means removing the additional position embedding matrix in Auto-embedding Blcok. (3) \textbf{\method{} w/o Staged-Adaptation} which means removing staged adaptation layers. (4) \textbf{\method{} w/ One-Adaptation-Layer} which means using only one adaptation layer for all the drafting heads. (5) \textbf{\method{} w/o Sampled-Token} which means removing sampled token during adaptation process. The experimental results are presented in Table \ref{tab:abla1}, \ref{tab:abla2}. From these comparative results, some observations can be found as follows:
\begin{itemize}[leftmargin=*]
    \item \textbf{\method{} w/o Auto-embedding} exhibits an approximate 5\%-8\% decrease in speed-up performance and about a 10\%-12\% reduction in average accepted length. This highlights the effectiveness of the Auto-embedding Block in mitigating inaccuracies deriving from the independent speculation of Medusa heads, and demonstrating the efficiency of non-autoregressive drafting computations. Additionally, \textbf{\method{} w/o Position-Encoding} exhibits a slight performance decline, with an approximate 2\% decrease in inference speed-up, suggesting that position encoding provides additional benefits.
    \item \textbf{\method{} w/o Staged-Adaptation} leads to a more significant decline in speed-up (14\%) and average accepted length (16\%). This emphasizes the importance of bridging the feature gap between the target model and drafting heads, and further underscores the critical role of the staged adaptation layer in enhancing the auto-embedding block. Additionally, it is noteworthy that \textbf{\method{} w/ One-Adaptation-Layer} utilizes only a single adaptation layer for all drafting positions. In contrast to staged adaptation, this approach poses greater challenges to the adaptation process, resulting in some performance degradation, thereby validating the rationale behind our staged adaptation design.
    \item \textbf{\method{} w/o Sampled-Token} also causes an approximate 8\% performance decline. Unlike previous works (e.g., Hydra), we do not use the sampled token directly for the next step of prediction. Instead, we adapt it along with the target model's hidden states. This not only indicates that the sampled token, in addition to target model hidden states, contains important semantic information, but also demonstrates the effectiveness of our staged adaptation approach.
    \item Thanks to the autoregressive characteristics and the substantial number of parameters in the MLP layers, Hydra exhibits great performance in average token length. However, the computational overhead of auto-regressive methods is huge, resulting in significant reductions when translated into final speed-up. In contrast, \method{} achieves a comparable average token length to Hydra, and due to the parallelism and efficiency of its non-autoregressive computations, it ultimately attains a more favorable overall trade-off.
\end{itemize}

\section{Related Work}\label{sec:rel}
Increasing techniques have been proposed to enhance the inference speed of large language models (LLMs), covering aspects of system hardware, model architecture, and decoding algorithms. A significant branch of these techniques is \textbf{Model Compression}, which includes methods such as model quantization \citep{yao2023comprehensive, dettmers2024qlora, liu2023llmqat, ma2024era}, pruning \citep{belcak2023exponentially, liu2023deja, zhong2024propd}, and distillation \citep{zhou2024distillspec, sun2024spectr, touvron2021training}. Additionally, techniques like kv-cache \citep{ge2023model, kwon2023efficient}, flash-attention \citep{dao2022flashattention}, and early exiting \citep{bae-etal-2023-fast, elhoushi2024layerskip, liu2024kangaroo} have also significantly reduced inference overhead.
Another important line is \textbf{Speculative Decoding}, which our work is based on. It can be broadly categorized into two types. The first treats the target model and draft model separately and independently, involving the use of a small language model \citep{kim2024speculative, leviathan2023fast, liu2024online, monea2023pass, chen2024sequoia, du2024glide}, external database, or n-grams pool \citep{he2024rest, fu2024break, kou2024cllms, ou2024lossless} to generate candidate token sequences or token trees \citep{miao2024specinfer}, which the LLM then verifies. The second type views the draft model as a dependent approximation of the target model, deriving the draft model directly from the target model or building additional modules on top of the target model for drafting \citep{stern2018blockwise, zhang2023draft, zhang2024recurrent, li2024eagle, cai2024medusa, kimexploring, xiao2024clover, ankner2024hydra}. Unlike these approaches, we propose a novel method using an auto-embedding block combined with staged adaptation layers to further enhance acceleration. 
\section{Conclusion}\label{sec:con}
We propose \method{}, an efficient non-autoregressive speculative decoding framework that accelerates inference through parallel decoding and improves alignment between target and draft models via feature adaptation. \method{} integrates two core components: the Auto-embedding Block, leveraging bi-directional self-attention for collaborative speculation among drafting heads, and Staged Adaptation Layers, transforming target model semantics for multi-step predictions. Additionally, \method{} exploits sampled tokens to further optimize alignment. Extensive experiments confirm the superiority of \method{}, showcasing the promise of non-autoregressive methods in speculative decoding.

\section*{Limitations}
While we have found and adhered to using bi-directional self-attention for non-autoregressive modeling as an efficient inference structure, we have not yet fully explored the optimal structure of the Auto-embedding Block module. Specifically, this includes experimenting with different intermediate sizes (i.e., the hidden dimensions used in self-attention computations) and increasing the number of self-attention layers within the auto-embedding block to enhance its modeling depth (see \ref{app:opt}). Both of these structural optimizations could potentially improve Amphista's acceleration performance within the current framework. Additionally, this work primarily focuses on scenarios where the batch size is equal to one, with limited optimization for larger batch sizes. We leave these areas as our future work and also hope that researchers interested in non-autoregressive inference acceleration will build upon this foundation.

\section*{Acknowledgement}
We acknowledge the helpful discussions from Kolorin Yan, Fuwei Yang, Ethan Yang, Xiandong Zhao, Mahdi Kamani, and Vikram Appia during the writing process of this work.

\bibliography{custom}

\clearpage
\appendix
\section{Appendix}

\subsection{Draft Tree} \label{app:draft}
\begin{figure}[t]
    \centering
    \includegraphics[width=\linewidth]{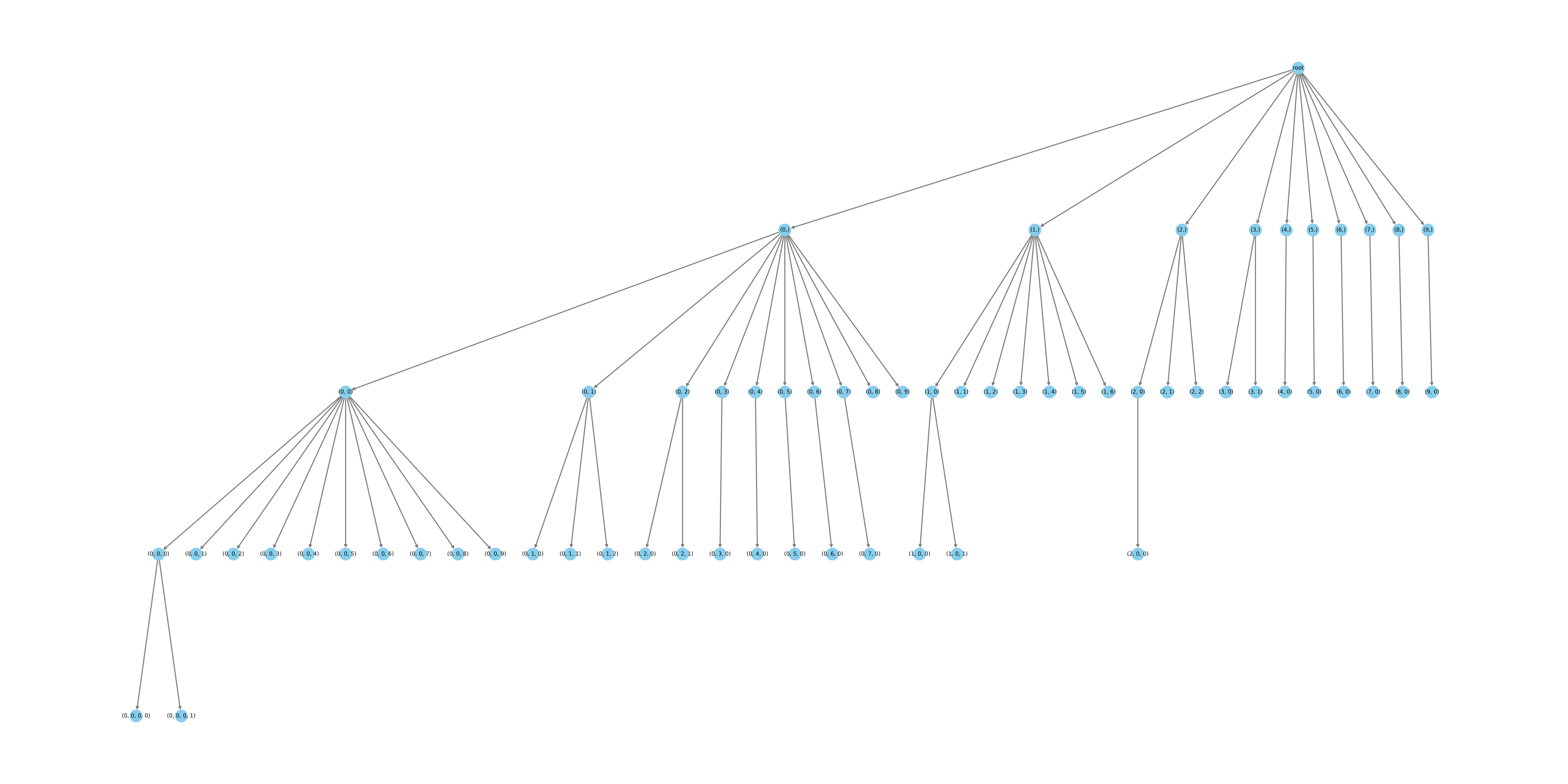}
    \caption{Draft tree used in Medusa, Hydra and our Amphista.}
    \label{fig:draft_tree}
\end{figure}

\begin{figure}[t]
    \centering
    \includegraphics[width=\linewidth]{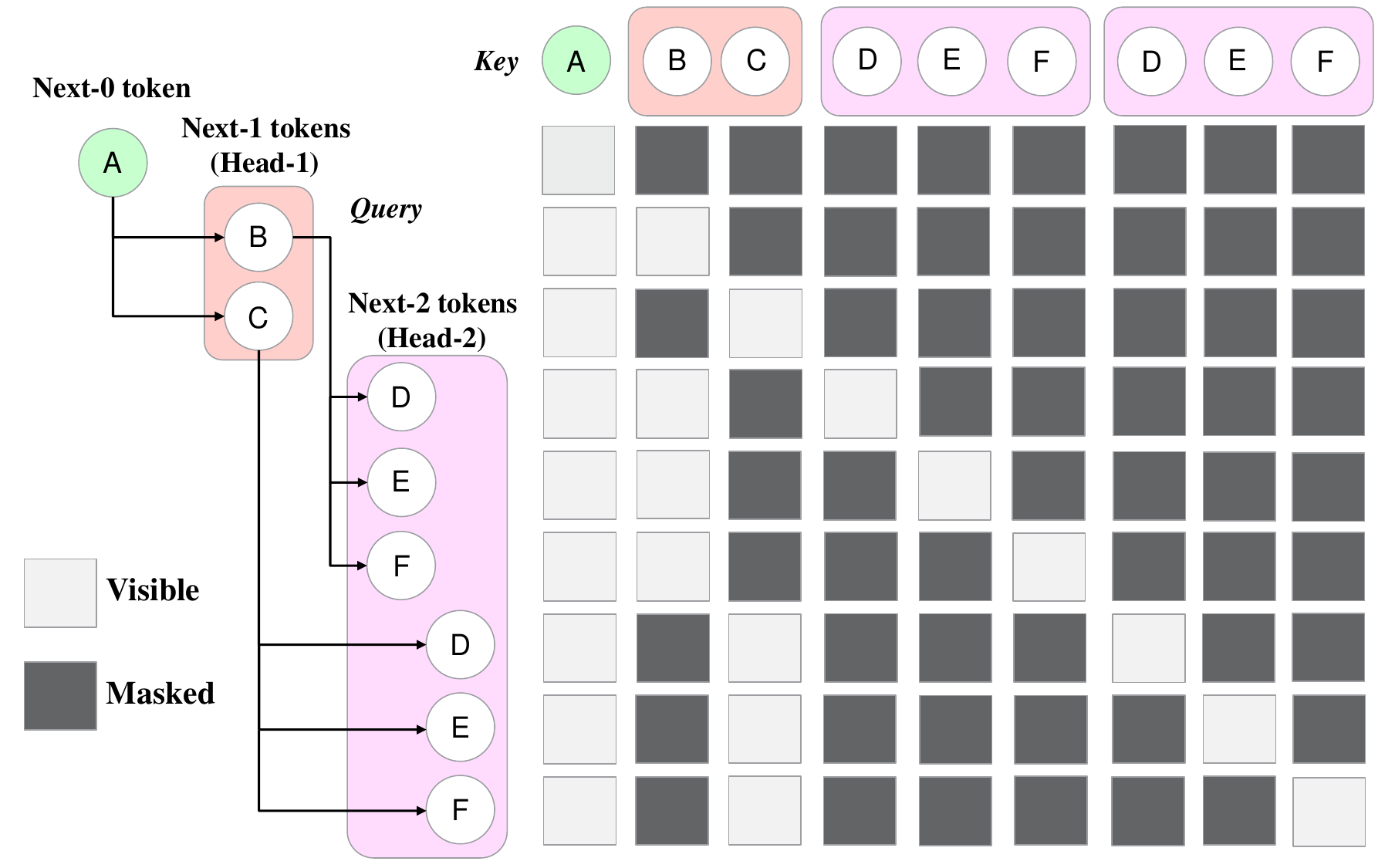}
    \caption{\textbf{An Illustration of Tree Attention.} Assuming Medusa has only 2 heads, where head-1 generates the top-2 tokens and head-2 generates the top-3 tokens, resulting in 6 candidate sequences (e.g., ABD). Additionally, a special tree mask is designed to ensure causal relationships among the top-k nodes of each head.}
    \label{fig:draft_mask}
\end{figure}

For a fully fair comparison, we adopt the same draft tree structure as Medusa and Hydra. As shown in Figure \ref{fig:draft_tree}, this tree is a sparse structure with a depth of 4, representing four drafting heads, and includes a total of 64 nodes, including the root node (the token sampled in the final step of the target model). Each layer's nodes represent the tokens obtained by top\_k sampling from the corresponding drafting head. The entire tree is constructed using an auxiliary dataset by maximizing the acceptance probability of the whole tree \cite{cai2024medusa}. Moreover, a specially designed tree mask is used to correctly compute attention scores while simultaneously handling multiple paths, as described in Figure \ref{fig:draft_mask}.

However, in some cases, due to the lack of redundant computational power (such as in high-throughput inference service scenarios) or parallel accelerators, an excessive number of tree nodes may lead to significant computation overhead, thereby affecting the acceleration efficiency of the algorithm. Consequently, we configure varying numbers of draft tree nodes without changing the tree depth for more comprehensive comparison, and the experimental results are shown in Table \ref{tab:tree_node}. From these results we observe that as the number of tree nodes decreases, the width of the tree reduces, leading to a decrease in speed-up for all compared methods. However, the decline is slightly less pronounced for Amphista, owing to its higher head accuracy. Furthermore, across various tree node configurations, we consistently achieve optimal performance, demonstrating the advantages of our algorithm in practical deployment and low-resource scenarios.

\begin{table}[h]
\centering
\caption{Speed-up comparison on MT-Bench with varying number of draft tree nodes.}
\label{tab:tree_node}
    \resizebox{0.5\textwidth}{!}{
\begin{tabular}{@{}l|cccc@{}}
\toprule
Method   & Node = 22  & Node = 35  & Node = 45 & Node = 64 \\ \midrule
Medusa   & 1.71$\times$     & 1.80$\times$      &  1.87$\times$       &  1.87$\times$         \\
Hydra++  & 2.17$\times$         & 2.26$\times$          &  2.28$\times$        &  2.37$\times$         \\
Amphista &\textbf{2.29}$\times$   & \textbf{2.37}$\times$         & \textbf{2.42}$\times$         &  \textbf{2.44}$\times$    \\ \bottomrule
\end{tabular}}
\end{table}

\subsection{Additional Experiments Results}\label{app:moreex}
\subsubsection{Evaluation on XSUM and CNN/DM} \label{app:xsum}
We use XSUM \cite{Narayan2018DontGM} and CNN/DM \cite{see-etal-2017-get} for evaluating the generation quality of Amphista, the target model is vicuna 7B. Specifically, we perform zero-shot evaluation and the input prompt template is \verb|'Article:'+ 'Original Text' + '\nSummary:'|. Additionally, for input prompts exceeding a length of 2048, we perform truncation to meet the target model's input requirements.

\begin{table}[h]
\centering
\caption{The speed-up metric comparison on Humaneval and GSM8K between different methods under greedy setting. The target model is vicuna 7B and 13B, and we regard the speed-up of vanilla auto-regressive decoding as 1.00$\times$.}
\label{tab:mathcode}
    \resizebox{0.5\textwidth}{!}{
\begin{tabular}{@{}clcccc@{}}
\toprule
\multicolumn{1}{l}{Model Size} & Benchmark & Vinilla AR & Medusa & Hydra++ & Amphista \\ \midrule
\multirow{2}{*}{7B}            & Humaneval & 1.00$\times$     &  2.40$\times$      & 2.76$\times$        & \textbf{3.02}$\times$          \\
                               & GSM8K     & 1.00$\times$     &   1.87$\times$     &  2.14$\times$      & \textbf{2.32}$\times$       \\ \midrule
\multirow{2}{*}{13B}           & Humaneval & 1.00$\times$       & 2.11$\times$      &  2.75$\times$       &  \textbf{3.00}$\times$       \\
                               & GSM8K     &  1.00$\times$          &  1.98$\times$      & 2.39$\times$        & \textbf{2.68}$\times$         \\ \bottomrule
\end{tabular}}
\end{table}

\begin{table*}[t]
\centering
\caption{The speed-up and average accepted length metric comparison with the target model being vicuna 7B. We regard the speed-up of vanilla auto-regressive decoding as 1.00$\times$.}
\label{tab:scaling}
    \resizebox{0.9\textwidth}{!}{
\begin{tabular}{ll|c|cccccc}
\toprule
\multirow{2}{*}{Metric} & \multirow{2}{*}{Method} & \multirow{2}{*}{MT-Bench} & \multicolumn{5}{c}{Spec-Bench} & \multirow{2}{*}{Avg} \\ \cline{4-8}
 &  &  & Translation & Summarization & QA & Math & RAG &  \\ \midrule
\multirow{4}{*}{Speed-up} 
& Vanilla & 1.00$\times$ & 1.00$\times$ & 1.00$\times$ & 1.00$\times$ & 1.00$\times$ & 1.00$\times$ & 1.00$\times$ \\ 
& Hydra++ & 2.37$\times$ & 1.92$\times$ & 1.80$\times$ & 1.94$\times$ & 2.43$\times$ & 2.04$\times$ & 2.03$\times$ \\
& EAGLE & 2.58$\times$ & 1.94$\times$ & 2.21$\times$ & 2.02$\times$ & 2.57$\times$ & 2.30$\times$ & 2.21$\times$ \\
 & \method{} & 2.44$\times$ & 1.96$\times$ & 2.11$\times$ & 1.94$\times$ & 2.45$\times$ & 2.20$\times$ & 2.13$\times$\\ 
  & \method{}-$\alpha$ & \textbf{2.63$\times$} & \textbf{2.09$\times$}  & \textbf{2.23$\times$} & \textbf{2.06$\times$} & \textbf{2.61$\times$} & \textbf{2.34$\times$} & \textbf{2.27$\times$} \\ \midrule
\multirow{4}{*}{Average Accepted Length} & Vanilla  & 1.00 & 1.00 & 1.00 & 1.00 & 1.00 & 1.00 & 1.00  \\
 & Hydra++ & 3.58 & 2.80 & 2.70 & 2.91 & 3.61 & 2.90 & 2.98\\
  & EAGLE & 3.84 & 2.92 & 3.32 & 3.14  & 3.93 & 3.31 & \textbf{3.32}\\
 & \method{}& 3.50 & 2.62 & 3.01 & 2.80  & 3.50 & 2.96 & 2.98 \\ 
 & \method{}-$\alpha$ & 3.58 & 2.70  & 3.14 & 2.90 & 3.62 & 3.08 & 3.09 \\ \midrule
\end{tabular}}
\end{table*}

\begin{table}[h]
\centering
\caption{Experiment results of LoRA-like lm heads optimization. Note that we consider the speed-up of full rank lm head as $1.00\times$.}
\label{tab:lora}
    \resizebox{0.5\textwidth}{!}{
\begin{tabular}{@{}l|ccccc@{}}
\toprule
Benchmark   & rank=4096 (full)  & rank=64  & rank=128 & rank=256 & rank=512\\ \midrule
MT-bench   & $1.00\times$     & $0.98\times$      &  $1.01\times$       &  $\mathbf{1.02\times}$  &  $0.99\times$       \\
Spec-bench & $1.00\times$   & $1.00\times$         & $1.00\times$         &  $\mathbf{1.01\times}$  & $0.98\times$    \\ \bottomrule
\end{tabular}}
\end{table}

\subsubsection{Code Generation and Math Reasoning}\label{app:code}
In this section, we provide more experimental results on code generation and math reasoning. we choose public Humaneval \cite{chen2021codex} and GSM8k \cite{cobbe2021gsm8k} benchmark for evaluation, and the target model is vicuna 7B and vicuna 13B. According to the results in Table \ref{tab:mathcode}, we can observe that due to the universal template and notation of code generation and mathematical reasoning, almost all compared methods achieve a higher speed-up. Furthermore, Amphista algorithm consistently attains optimal performance, demonstrating the superiority of our approach.

\subsubsection{Exploring The Potential of Amphista}\label{app:opt}
In this section, we conduct a preliminary exploration of \method{}'s scaling ability to demonstrate its potential for performance enhancement. By leveraging the efficiency of non-autoregressive modeling, we increase the number of auto-embedding blocks, which are essential modules within \method{}, while maintaining parallel inference. This approach yields remarkable results, detailed in Table \ref{tab:scaling}. Specifically, we employ two layers of self-attention in the auto-embedding module, renaming our method as \method{}-$\alpha$. This adjustment leads to an average accepted length increase of approximately 0.1-0.2 tokens and a notable 5\%-8\% improvement in overall speed-up, highlighting \method{}'s performance growth potential. We anticipate this to be a highly promising and potent attribute of Amphista.

\subsubsection{Parameter Complexity Optimization of Amphista}\label{app:plex}
In this part, we propose LoRA-like drafting lm heads to further optimize the original learnable lm heads of Amphista, which significantly reduces the parameter amount and complexity. Specifically, we use two low-rank matrices to replace the original lm head matrix. The experimental results are shown in Table \ref{tab:lora}, we choose Vicuna 7B as target model, so the parameter count of lm head is 4096 * 32000. With the increase of rank, we can reduce the number of learnable parameters by up to 45\% while maintaining almost the same performance, which greatly reduces the complexity of model parameters and reflects the advantages and potential of \method{} in practical deployment.

\begin{figure}[t]
    \centering
    \includegraphics[width=\linewidth]{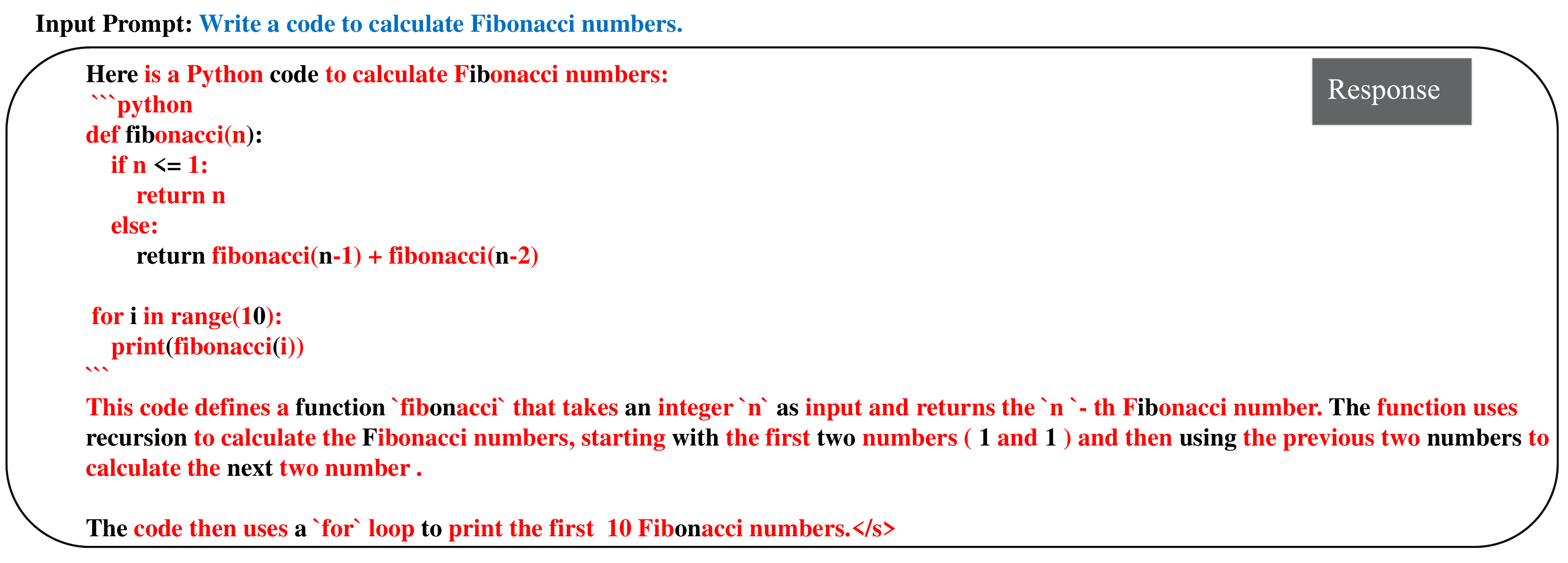}
    \caption{\textbf{Case Study on Code Generation}. Tokens in \textbf{\textcolor{red}{red}} means those generated by Amphista and tokens in \textbf{\textcolor{black}{black}} means those generated by target model itself.}
    \label{fig:case-1}
\end{figure}

\begin{figure}[t]
    \centering
    \includegraphics[width=\linewidth]{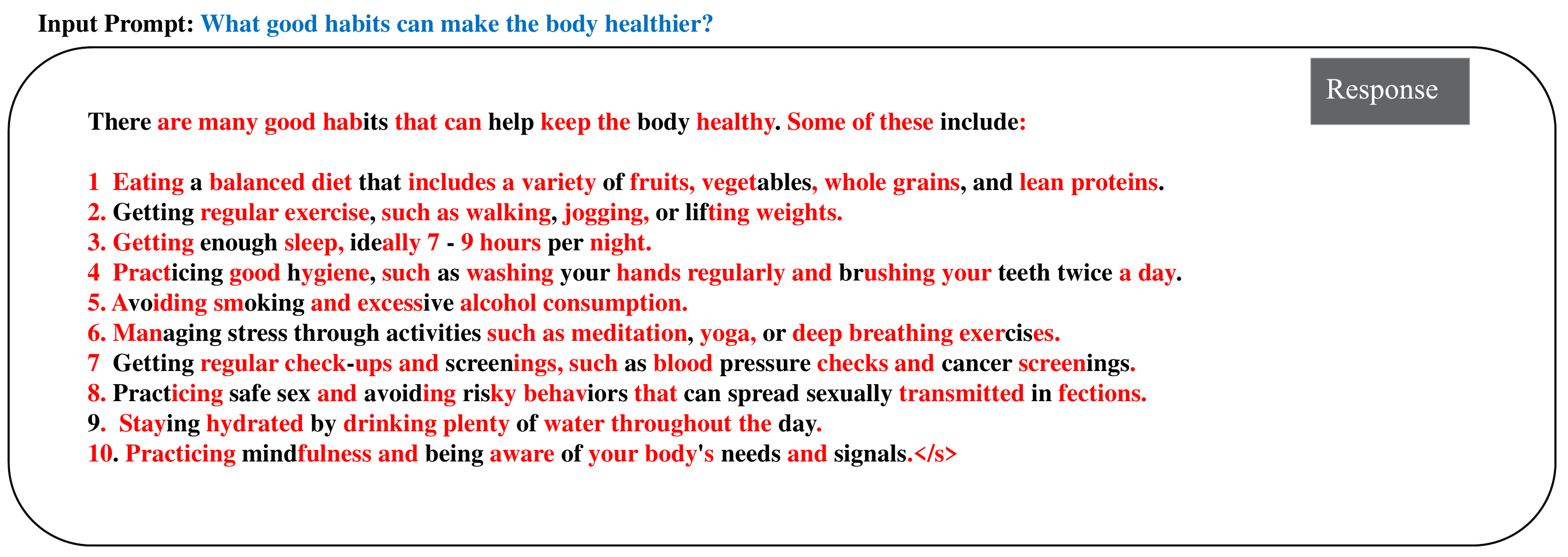}
    \caption{\textbf{Case Study on Text Generation.} Tokens in \textbf{\textcolor{red}{red}} means those generated by Amphista and tokens in  \textbf{\textcolor{black}{black}} means those generated by target model itself.}
    \label{fig:case-2}
\end{figure}

\subsection{Case Study}
Here we show some real case studies (see Figure \ref{fig:case-1}, \ref{fig:case-2}) on Amphista inference, the target model is Vicuna 7B. Note that we do not apply any special processing to the tokenizer's output, preserving the original results. Tokens highlighted in \textbf{red} represent those generated by Amphista during each step of decoding. Tokens in \textbf{black} indicate those generated by target model. From these practical examples, we can observe that in the vast majority of cases, Amphista generates at least two tokens per decoding step. This generally results in a stable at least 2x speed-up, demonstrating the efficiency of our algorithm. Additionally, Amphista's output is consistent with the target model's auto-regressive decoding output, ensuring the generation quality of Amphista.

\end{document}